\documentclass{article}

    \usepackage[final]{neurips_2021}

\usepackage{xr-hyper}
\usepackage[utf8]{inputenc} 
\usepackage[T1]{fontenc}    
\usepackage{hyperref}       
\usepackage{url}            
\usepackage{booktabs}       
\usepackage{amsfonts}       
\usepackage{nicefrac}       
\usepackage{microtype}      
\usepackage{xcolor}         
\usepackage{graphicx}
\usepackage{caption}
\usepackage{subcaption}
\usepackage{fancyvrb}
\usepackage{color}
\usepackage{code_format}
\usepackage{amsmath,amsthm,mathtools,amssymb}
\usepackage{tikz}
\usepackage{soul}
\usepackage{natbib}
\usepackage{hyperref}
\DeclareUnicodeCharacter{2212}{-}

\usetikzlibrary{graphs,graphs.standard,shapes,arrows.meta,calc}

\usepackage[nameinlink]{cleveref}
\renewcommand{\autoref}{\Cref}

\newtheorem{theorem}{Theorem}

\usepackage{mathtools}
\usepackage{xcolor}

\theoremstyle{definition}
\newtheorem{definition}{Definition}

\newcommand{\R}{\mathbb{R}}

\newcommand{\states}{\mathcal{S}}
\newcommand{\aN}{N}
\newcommand{\agentset}{[\aN]}
\newcommand{\actions}{\mathcal{A}}

\newcommand{\transition}{P}
\newcommand{\reward}{R}
\newcommand{\observations}{\Omega}
\newcommand{\obsfunc}{O}
\newcommand{\observation}{\omega}
\DeclarePairedDelimiter{\set}{\{}{\}}

\newcommand{\agtrans}{\nu}
\newcommand{\dettrans}{T}
\newcommand{\agentenvset}{\mathcal{N}}

\newcommand{\rewardspace}{\mathcal{R}}

\definecolor{agent1}{RGB}{176,99,254}
\definecolor{agent2}{RGB}{15,102,170}
\definecolor{river}{RGB}{99,156,194}
\definecolor{cleaningbeam}{RGB}{100,255,255}
\definecolor{waste}{RGB}{113,75,24}
\newcommand{\agentturn}{\tau}

\newcommand{\infopart}{H}

\newcommand{\actseq}{\tilde{\actions}}
\newcommand{\terminals}{T}

\def\sz{0.45cm}

\title{PettingZoo: A Standard API for Multi-Agent Reinforcement Learning}

\author{J. K. Terry\thanks{Swarm Labs} \ \thanks{Department of Computer Science | University of Maryland, College Park}\\
\texttt{j.k.terry@swarmlabs.com}\\
\And
Benjamin Black\footnotemark[1] \ \footnotemark[2]\\
\texttt{benjamin.black@swarmlabs.com}\\
\And
Nathaniel Grammel\footnotemark[2]\\
\texttt{ngrammel@umd.edu}
\And
Mario Jayakumar\footnotemark[2]\\
\texttt{mariojay@umd.edu}\\
\And
Ananth Hari\thanks{Department of Electrical and Computer Engineering | University of Maryland, College Park}\\
\texttt{ahari1@umd.edu}\\
\And
Ryan Sullivan\footnotemark[1] \ \footnotemark[2]\\
\texttt{ryan.sullivan@swarmlabs.com}\\
\And
Luis Santos\thanks{Department of Mechanical Engineering | University of Maryland, College Park}\\
\texttt{lss@umd.edu}\\
\And
Rodrigo Perez \ \thanks{Maryland Robotics Center | University of Maryland, College Park}\\
\texttt{rlazcano@umd.edu}\\
\And
Caroline Horsch\footnotemark[1] \ \footnotemark[2]\\
\texttt{caroline.horsch@swarmlabs.com}\\
\And
Clemens Dieffendahl \thanks{Faculty of Electrical Engineering and Computer Science | Technical University of Berlin}\\
\texttt{dieffendahl@campus.tu-berlin.de}\\
\And
Niall L. Williams\footnotemark[2]\\
\texttt{niallw@umd.edu}\\
\And
Yashas Lokesh\footnotemark[2]\\
\texttt{yashloke@umd.edu}\\
\And
Praveen Ravi\footnotemark[2]\\
\texttt{pravi@umd.edu}\\
}

\begin{document}

\maketitle

\begin{abstract}
This paper introduces the PettingZoo library and the accompanying Agent Environment Cycle (``AEC'') games model. PettingZoo is a library of diverse sets of multi-agent environments with a universal, elegant Python API. PettingZoo was developed with the goal of accelerating research in Multi-Agent Reinforcement Learning (``MARL''), by making work more interchangeable, accessible and reproducible akin to what OpenAI's Gym library did for single-agent reinforcement learning. PettingZoo's API, while inheriting many features of Gym, is unique amongst MARL APIs in that it's based around the novel AEC games model. We argue, in part through case studies on major problems in popular MARL environments, that the popular game models are poor conceptual models of games commonly used in MARL and accordingly can promote confusing bugs that are hard to detect, and that the AEC games model addresses these problems.


\end{abstract}

\section{Introduction}
Multi-Agent Reinforcement Learning (MARL) has been behind many of the most publicized achievements of modern machine learning --- AlphaGo Zero \citep{silver2017mastering}, OpenAI Five \citep{OpenAI_dota}, AlphaStar \citep{vinyals2019grandmaster}. These achievements motivated a boom in MARL research, with Google Scholar indexing 9,480 new papers discussing multi-agent reinforcement learning in 2020 alone. Despite this boom, conducting research in MARL remains a significant engineering challenge. A large part of this is because, unlike single agent reinforcement learning which has OpenAI's Gym, no de facto standard API exists in MARL for how agents interface with environments. This makes the reuse of existing learning code for new purposes require substantial effort, consuming researchers' time and preventing more thorough comparisons in research. This lack of a standardized API has also prevented the proliferation of learning libraries in MARL. While a massive number of Gym-based single-agent reinforcement learning libraries or code bases exist (as a rough measure 669 pip-installable packages depend on it at the time of writing \cite{openai}), only 5 MARL libraries with large user bases exist \citep{openspiel2019,tianshou,liang2018rllib,smac2019, nota2020autonomous}. The proliferation of these Gym based learning libraries has proved essential to the adoption of applied RL in fields like robotics or finance and without them the growth of applied MARL is a significantly greater challenge. Motivated by this, this paper introduces the PettingZoo library and API, which was created with the goal of making research in MARL more accessible and serving as a multi-agent version of Gym.







Prior to PettingZoo, the numerous single-use MARL APIs almost exclusively inherited their design from the two most prominent mathematical models of games in the MARL literature---Partially Observable Stochastic Games (``POSGs'') and Extensive Form Games (``EFGs''). During our development, we discovered that these common models of games are not conceptually clear for multi-agent games implemented in code and cannot form the basis of APIs that cleanly handle all types of multi-agent environments.

To solve this, we introduce a new formal model of games, Agent Environment Cycle (``AEC'') games that serves as the basis of the PettingZoo API. We argue that this model is a better conceptual fit for games implemented in code. and is uniquely suitable for general MARL APIs. We then prove that any AEC game can be represented by the standard POSG model, and that any POSG can be represented by an AEC game. To illustrate the importance of the AEC games model, this paper further covers two case studies of meaningful bugs in popular MARL implementations. In both cases, these bugs went unnoticed for a long time. Both stemmed from using confusing models of games, and would have been made impossible by using an AEC games based API.

The PettingZoo library can be installed via \texttt{pip install pettingzoo}, the documentation is available at \url{https://www.pettingzoo.ml}, and the repository is available at \url{https://github.com/Farama-Foundation/PettingZoo}.

\section{Background and Related Works}
\label{sec:state_apis}
Here we briefly survey the state of modeling and APIs in MARL, beginning by briefly looking at Gym's API (\autoref{fig:gym_example}). This API is the de facto standard in single agent reinforcement learning, has largely served as the basis for subsequent multi-agent APIs, and will be compared to later.

\begin{figure}[ht]
\centering
\begin{minipage}{.52\linewidth}
\begin{center}
\begin{BVerbatim}[commandchars=\\\{\},fontsize=\scriptsize]
\PY{k+kn}{import} \PY{n+nn}{gym}
\PY{n}{env} \PY{o}{=} \PY{n}{gym}\PY{o}{.}\PY{n}{make}\PY{p}{(}\PY{l+s+s1}{\PYZsq{}}\PY{l+s+s1}{CartPole\PYZhy{}v0}\PY{l+s+s1}{\PYZsq{}}\PY{p}{)}
\PY{n}{observation} \PY{o}{=} \PY{n}{env}\PY{o}{.}\PY{n}{reset}\PY{p}{(}\PY{p}{)}
\PY{k}{for} \PY{n}{\PYZus{}} \PY{o+ow}{in} \PY{n+nb}{range}\PY{p}{(}\PY{l+m+mi}{1000}\PY{p}{)}\PY{p}{:}
    \PY{n}{action} \PY{o}{=} \PY{n}{policy}\PY{p}{(}\PY{n}{observation}\PY{p}{)}
    \PY{n}{observation}\PY{p}{,} \PY{n}{reward}\PY{p}{,} \PY{n}{done}\PY{p}{,} \PY{n}{info} \PY{o}{=} \PY{n}{env}\PY{o}{.}\PY{n}{step}\PY{p}{(}\PY{n}{action}\PY{p}{)}
\end{BVerbatim}
\caption{An example of the basic usage of Gym}
\label{fig:gym_example}
\end{center}
\end{minipage}%
\hspace{0.02\linewidth}
\begin{minipage}{0.45\linewidth}
\begin{center}
\begin{BVerbatim}[commandchars=\\\{\},fontsize=\scriptsize]
\PY{k}{from} \PY{n+nn}{ray}\PY{n}{.}\PY{n+nn}{rllib}\PY{n}{.}\PY{n+nn}{examples}\PY{n}{.}\PY{n+nn}{env}\PY{n}{.}\PY{n+nn}{multi_agent}
    \PY{k+kn}{import} \PY{n}{MultiAgentCartPole}
\PY{n}{env} \PY{o}{=} \PY{n}{MultiAgentCartPole}\PY{p}{(}\PY{p}{)}
\PY{n}{observation} \PY{o}{=} \PY{n}{env}\PY{o}{.}\PY{n}{reset}\PY{p}{(}\PY{p}{)}
\PY{k}{for} \PY{n}{\PYZus{}} \PY{o+ow}{in} \PY{n+nb}{range}\PY{p}{(}\PY{l+m+mi}{1000}\PY{p}{)}\PY{p}{:}
    \PY{n}{actions} \PY{o}{=} \PY{n}{policies}\PY{p}{(}\PY{n}{agents, observation}\PY{p}{)}
    \PY{n}{observation}\PY{p}{,} \PY{n}{rewards}\PY{p}{,} \PY{n}{dones}\PY{p}{,}
        \PY{n}{infos} \PY{o}{=} \PY{n}{env}\PY{o}{.}\PY{n}{step}\PY{p}{(}\PY{n}{actions}\PY{p}{)}
\end{BVerbatim}
\caption{An example of the basic usage of RLlib}
\label{fig:rllib_example}
\end{center}
\end{minipage}
\end{figure}
The Gym API is a fairly straightforward Python API that borrows from the POMDP conceptualization of RL. The API's simplicity and conceptual clarity has made it highly influential, and it naturally accompanying the pervasive POMDP model that's used as the pervasive mental and mathematical model of reinforcement learning~\citep{brockman2016openai}. This makes it easier for anyone with an understanding of the RL framework to understand Gym's API in full. 

\subsection{Partially Observable Stochastic Games and RLlib} \label{sec:posg-rllib}

Multi-agent reinforcement learning does not have a universal mental and mathematical model like the POMDP model in single-agent reinforcement learning. One of the most popular models is the partially observable stochastic game (``POSG''). This model is very similar to, and strictly more general than, multi-agent MDPs~\citep{boutilier1996planning}, Dec-POMDPs~\citep{bernstein2002DecPOMDP}, and Stochastic (``Markov'') games~\citep{shapley1953stochasticgames}). In a POSG, all agents step together, observe together, and are rewarded together. The full formal definition is presented in \Cref{sec:posgdef}

This model of simultaneous stepping naturally translates into Gym-like APIs, where the actions, observations, rewards, and so on are lists or dictionaries of individual values for agents. This design choice has become the standard for MARL outside of strictly turn-based games like poker, where simultaneous stepping would be a poor conceptual fit \citep{lowe2017multi,zheng2017magent,gupta2017cooperative,deepmind_soccer,liang2018rllib, tianshou}. One example of this is shown in \autoref{fig:rllib_example} with the multi-agent API in RLlib \citep{liang2018rllib}, where agent-keyed dictionaries of actions, observations and rewards are passed in a simple extension of the Gym API.


This model has made it much easier to apply single agent RL methods to multi-agent settings. However, there are two immediate problems with this model:

\begin{enumerate}
    \item Supporting strictly turn-based games like chess requires constantly passing dummy actions for non-acting agents (or using similar tricks).
    \item Changing the number of agents for agent death or creation is very awkward, as learning code has to cope with lists suddenly changing sizes.
\end{enumerate}

\subsection{OpenSpiel and Extensive Form Games} \label{sec:efg-openspiel}

In the cases of strictly turn based games where POSG models are poorly suited (e.g. Chess), MARL researchers generally mathematically model the games as Extensive Form Games (``EFG''). The EFG represents games as a tree, \emph{explicitly} representing every possible sequence of actions as a root to leaf path in the tree. Stochastic aspects of a game (or MARL environment) are captured by adding a ``Nature'' player (sometimes also called ``Chance'') which takes actions according to some given probability distribution. For a full definition of EFGs, we refer the reader to~\citet{osborne1994gametheory} or \autoref{sec:efgdef}.
OpenSpiel \citep{openspiel2019}, a major library with a large collection of classical board and card games for MARL bases their API off of the EFG paradigm, the API of which is shown in \autoref{fig:openspiel_example}.



\begin{figure}[ht]
\begin{center}
\begin{BVerbatim}[commandchars=\\\{\},fontsize=\scriptsize]

\PY{k+kn}{import} \PY{n+nn}{pyspiel}
\PY{k+kn}{import} \PY{n+nn}{numpy} \PY{k}{as} \PY{n+nn}{np}

\PY{n}{game} \PY{o}{=} \PY{n}{pyspiel}\PY{o}{.}\PY{n}{load\PYZus{}game}\PY{p}{(}\PY{l+s+s2}{\PYZdq{}}\PY{l+s+s2}{kuhn\PYZus{}poker}\PY{l+s+s2}{\PYZdq{}}\PY{p}{)}
\PY{n}{state} \PY{o}{=} \PY{n}{game}\PY{o}{.}\PY{n}{new\PYZus{}initial\PYZus{}state}\PY{p}{(}\PY{p}{)}
\PY{k}{while} \PY{o+ow}{not} \PY{n}{state}\PY{o}{.}\PY{n}{is\PYZus{}terminal}\PY{p}{(}\PY{p}{)}\PY{p}{:}
  \PY{k}{if} \PY{n}{state}\PY{o}{.}\PY{n}{is\PYZus{}chance\PYZus{}node}\PY{p}{(}\PY{p}{)}\PY{p}{:}
    \PY{c+c1}{\PYZsh{} Step the stochastic environment.}
    \PY{n}{action\PYZus{}list}\PY{p}{,} \PY{n}{prob\PYZus{}list} \PY{o}{=} \PY{n+nb}{zip}\PY{p}{(}\PY{o}{*}\PY{n}{state}\PY{o}{.}\PY{n}{chance\PYZus{}outcomes}\PY{p}{(}\PY{p}{)}\PY{p}{)}
    \PY{n}{state}\PY{o}{.}\PY{n}{apply\PYZus{}action}\PY{p}{(}\PY{n}{np}\PY{o}{.}\PY{n}{random}\PY{o}{.}\PY{n}{choice}\PY{p}{(}\PY{n}{action\PYZus{}list}\PY{p}{,} \PY{n}{p}\PY{o}{=}\PY{n}{prob\PYZus{}list}\PY{p}{)}\PY{p}{)}
  \PY{k}{else}\PY{p}{:}
    \PY{c+c1}{\PYZsh{} sample an action for the agent}
    \PY{n}{legal\PYZus{}actions} \PY{o}{=} \PY{n}{state}\PY{o}{.}\PY{n}{legal\PYZus{}actions}\PY{p}{(}\PY{p}{)}
    \PY{n}{observations} \PY{o}{=} \PY{n}{state}\PY{o}{.}\PY{n}{observation\PYZus{}tensor}\PY{p}{(}\PY{p}{)}
    \PY{n}{action} \PY{o}{=} \PY{n}{policies}\PY{p}{(}\PY{n}{state}\PY{o}{.}\PY{n}{current\PYZus{}agent}\PY{p}{(}\PY{p}{)}\PY{p}{,} \PY{n}{legal\PYZus{}actions}\PY{p}{,} \PY{n}{observations}\PY{p}{)}
    \PY{n}{state}\PY{o}{.}\PY{n}{apply\PYZus{}action}\PY{p}{(}\PY{n}{action}\PY{p}{)}
    \PY{n}{rewards} \PY{o}{=} \PY{n}{state}\PY{o}{.}\PY{n}{rewards}\PY{p}{(}\PY{p}{)}

\end{BVerbatim}
\caption{An example of the basic usage of OpenSpiel}
\label{fig:openspiel_example}
\end{center}
\end{figure}

The EFG model has been successfully used for solving problems involving theory of mind with methods like game theoretic analysis and tree search. However, for application in general MARL problems, three immediate concerns arise with the EFG model:


\begin{enumerate}
  \item The model, and the corresponding API, is very complex compared to that of POSGs, and isn't suitable for beginners the way Gym is---this environment API is much more complicated than Gym's API or RLLib's POSG API for example. Furthermore, due to the complexity of the EFG model, reinforcement learning researchers don't ubiquitously use it as a mental model of games in the same way that they use the POSG or POMDP model.
  \item The formal definition only includes rewards at the end of games, while reinforcement learning often requires frequent rewards. While this is possible to work around in the API implementation, it is not ideal.
  \item The OpenSpiel API does not handle continuous actions (a common and important case in RL), though this was a choice that is not inherent to the EFG model.
\end{enumerate}

It's also worth briefly noting that some simple strictly turn based games are modeled with the single-agent Gym API, with the environment alternating which agent is controlled, \citep{slimevolleygym}. This approach is unable to reasonably scale beyond two agents due to the difficulties of handling changes in agent order (e.g. Uno), agent death, and agent creation.

\section{PettingZoo Design Goals}

Our development of PettingZoo both as a general library and an API centered around the following goals.

\subsection{Be like Gym}

In PettingZoo, we wanted to leverage Gym's ubiquity, simplicity and universality. This created two concrete goals for us:

\begin{itemize}
  \item Make the API look and feel like Gym, and relatedly make the API pythonic and simple
  \item Include numerous reference implementations of games with the main package
\end{itemize}

Reusing as many design metaphors from Gym as possible will help its massive existing user base to almost instantly understand PettingZoo's API. Similarly, for an API to become standardized, it must support a large collection of useful environments to attract users and for adoption to begin, similar to what Gym did.



\subsection{Be a Universal API}

If there is to be a Gym-like API for MARL, it has to be able to support all use cases and types of environments. Accordingly, several technically difficult cases exist that have to be carefully considered:

\begin{itemize}
  \item Environments with large numbers of agents
  \item Environments with agent death and creation
  \item Environments where different agents can be chosen to participate in each episode
  \item Learning methods that require access to specialty low level features
\end{itemize}

Two related softer design goals for universal design are ensuring the API is simple enough for beginners to easily use, and making the API easily changeable if the direction of research in the field dramatically changes.





\section{Case Studies of Problems With The  POSG Model in MARL}
\label{sec:posgproblems}

To supplement the description of the problems with the POSG models described in \autoref{sec:posg-rllib},  we overview problems with basing APIs around these models that could theoretically occur in software games, and then examine real cases of those problems occurring in popular MARL environments. We specifically focus on POSGs here because EFG based APIs are extraordinarily rare (OpenSpiel is the only major one), while POSG based ones are almost universal.


\subsection{POSGs Don't Allow Access To Information You Should Have}
\label{sec:case_study_pursuit}

Another problem with modeling environments using simultaneous actions in the POSG model is that all of an agent's rewards (from all sources) are summed and returned all at once. In a multi-agent game though, this combined reward is often the composite reward from the actions of other agentss and the environment. Similarly, you might want to be able to attribute the source of this reward for various learning reasons, or for debugging purposes to find out the origin of your rewards. However, in thinking about reward origins, having all rewards emitted at once proves to be very confusing because rewards from different sources are all combined. Accessing this information via an API modeled after a POSG requires deviating from the model. This would come in the form of returning a 2D array of rewards instead of a list, which would be difficult to standardize and inconvenient for learning code to parse.

\begin{figure}[ht]
\centering
\includegraphics[width=.22\linewidth]{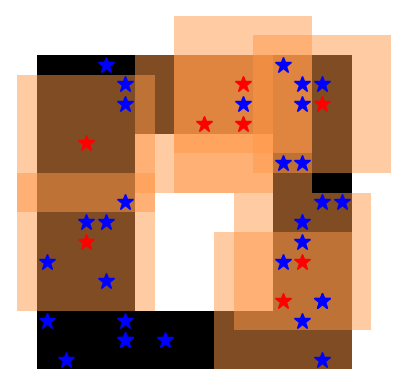}
\caption{The \emph{pursuit} environment from \citet{gupta2017cooperative}.}
\label{fig:pursuit}
\end{figure}

A notable case where this caused an issue in practice is in the popular pursuit gridworld  environment from \citet{gupta2017cooperative}, shown in \autoref{fig:pursuit}. In it, $8$ red controllable pursuer must work together to surround and capture $30$ randomly moving blue evaders. The action space of each pursuer is discrete (cardinal directions or do nothing), and the observation space is a $7 \times 7$ box centered around a pursuer (depicted by the orange box). When an evader is surrounded on all sides by pursuers or the game boundaries, each contributing pursuer gets a reward of $5$. 

In pursuit, pursuers move first, and then evaders move randomly, before it's determined if an evader is captured and rewards are emitted. Thus an evader that ``should have'' been captured is not actually captured. Having the evaders move second isn't a bug, it's just way of adding complexity to the classic genre of pursuer/evader multi-agent environments~\citep{vidal2002probabilistic}, and is representative of real problems. When \emph{pursuit} is viewed as an AEC game, we're forced to attribute rewards to individual steps, and the breakdown becomes pursuers receiving deterministic rewards from surrounding the evader, and then random reward due to the evader moving after. Removing this random component of the reward (the part caused by the evaders action after the pursuers had already moved), should then lead to superior performance. In this case the problem was so innocuous that fixing it required switching two lines of code where their order made no obvious difference. We experimentally validate this performance improvement in \autoref{sec:appendix_bug_ssd}, showing that on average this change resulted in up to a 22\% performance in the expected reward of a learned policy.

Bugs of this family could easily happen in almost any MARL environment, and analyzing and preventing them is made much easier when using the POSG model. Because every agent's rewards are summed together in the POSG model, this specific problem when looking at the code was extraordinarily non-obvious, whereas when forced to attribute the reward of individual agents this becomes clear. Moreover if an existing environment had this problem, by exposing the actual sources of rewards to learning code researchers are able to remove differing sources of reward to more easily find and remove bugs like this, and in principle learning algorithms could be developed that automatically differently weighted different sources of reward.

\subsection{POSGs Based APIs Are Not Conceptually Clear For Games Implemented In Code}
\label{sec:case_study_ssd}

Introducing race conditions is a very easy mistake to make in MARL code in practice, and this occurs because simultaneous models of multi-agent games are not representative of how game code normally executes. This stems from a very common scenario in multi-agent environments where two agents are able to take conflicting actions (i.e.~moving into the same space). This discrepancy has to be resolved by the environment (i.e. collision handling); which we call ``tie-breaking.'' 

Consider an environment with two agents, Alice and Bob, in which Alice steps first and tie-breaking is biased in Alice's favor. If such an environment were assumed to have simultaneous actions, then observations for both agents would be taken before either acted, causing the observation Bob acts on to no longer be an accurate representation of the environment if a conflict with biased tie-breaking occurs. For example, if both agents tried to step into the same square and Alice got the square because she was first in the list, Bob's observation before acting was effectively inaccurate and the environment was not truly parallel. This behavior is a true race condition---the result of stepping through the environment can inadvertently differ depending on the internal resolution order of agent actions. 

In any environment that's even slightly complex, a tremendous number of instances where tie-breaking must be handled will typically occur. In any cases where a single one is missed, the environment will have race conditions that your code will attempt to learn. While finding these will always be important, a valuable tool to mitigate these possibilities is to use an API that treats each agent as acting sequentially, returning new observations afterwards. This entirely prevents the opportunity for introducing race conditions. Moreover, this entire problem stems from the fact that using APIs that model agents as updating sequentially for software MARL environments generally makes more conceptual sense than modeling the updates as simultaneous---unless the authors of environments use very complex parallelization, the environments will \emph{actually} be updated one agent at a time. It is worth mentioning that this race condition cannot occur in an environment simulated in the physical world with continuous time or a simulated environment with a sufficient amount of observation delay (though most actively researched environment in MARL do not currently have any observation delay).

In \autoref{sec:appendix_bug_ssd} we go through a case study of a race condition like this happening in the open source implementation of the social sequential dilemma game environments~\citep{SSDOpenSource}. These are popular multi-agent grid world environments intended to study emergent behaviors for various forms of resource management, and has imperfect tie-breaking in a case where two agents try to act on resources in the same grid while using a simultaneous API. This bug in particular illustrates how extraordinarily difficult making all tie-breaking truly unbiased is in practice even for fairly simple environments. We defer this to the appendix as explaining the specific origin requires a large amount of exposition and diagrams about the rules of the environment.

\section{The Agent Environment Cycle Games Model}
\label{aec_model}

Motivated by the problems with applying the POSG and EFG models to MARL APIs, we developed the Agent Environment Cycle (``AEC'') Game. In this model, agents sequentially see their observation, agents take actions, rewards are emitted from the other agents, and the next agent to act is chosen. This is effectively a sequentially stepping form of the POSG model. 

Modeling multi-agent environments sequentially for APIs has numerous benefits:

\begin{itemize}
  \item It allows for clearer attribution of rewards to different origins, allowing for various learning improvements, as described in \autoref{sec:case_study_pursuit}.
  \item It prevents developers adding confusing and easy-to-introduce race conditions, as described in \autoref{sec:case_study_ssd}.
  \item It more closely models how computer games are executed in code, as described in \autoref{sec:case_study_ssd}.
  \item It formally allows for rewards after every step as is required in RL, but is not generally a part of the EFG model, as discussed in \autoref{sec:efg-openspiel}.
  \item It is simple enough to serve as a mental model, especially for beginners, unlike the EFG model as discussed in \autoref{sec:efg-openspiel}  and illustrated in the definition in \autoref{sec:efgdef}.
  \item Changing the number of agents for agent death or creation is less awkward, as learning code does not have to account for lists constantly changing sizes, as discussed in \autoref{sec:posg-rllib}.
  \item It is the least bad option for a universal API, compared to simultaneous stepping, as alluded to in \autoref{sec:posg-rllib}. Simultaneous stepping requires the use of no-op actions if not all agents can act which are very difficult to deal with, whereas sequentially stepping agents that could all act simultaneously and queuing up their actions is not especially inconvenient.
\end{itemize}

In \autoref{sec:aecdef} we mathematically formalize the AEC games model, however understanding the formalism in full is not essential to understanding the paper. In \autoref{sec:omitted_proofs} we further prove that for every AEC game an equivalent POSG exists and that for every POSG an equivalent AEC game exists. This shows that the AEC games model is as powerful a model as the most common current model of multi-agent environments.

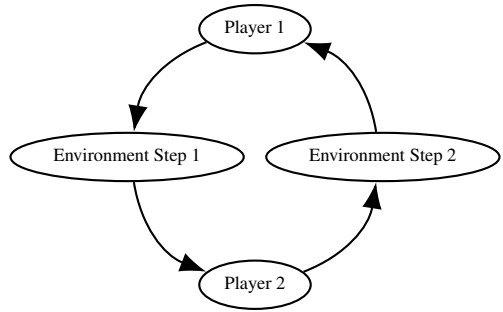
\begin{figure*}[ht]
    \centering
        \centering
        \begin{tikzpicture}[thick,>={Latex[length=.35cm]},font=\scriptsize]
            \graph[nodes={draw,ellipse}, counterclockwise, radius=1.7cm] {
                subgraph C_n[->, V={Player 1, Environment Step 1, Player 2, Environment Step 2}, edges={bend right=30}];
            };
        \end{tikzpicture}
        \caption{The AEC diagram of Chess}
        \label{fig:pong_aec-1}
\end{figure*}

One additional conceptual feature of the AEC games model exists that we have not previously discussed because it does not usually play a role in APIs (see \autoref{sec:additional_api}). In the AEC games model, we deviate from the POSG model by introducing the ``environment'' agent, which is analogous to the Nature agent from EFGs. When this agent acts in the model it indicates the updating of the environment itself, realizing and reacting to submitting agent actions. This allows for a more comprehensive attribution of rewards, causes of agent death, and discussion of games with strange updating rules and race conditions. An example of the transitions for Chess is shown in \autoref{fig:pong_aec-1}, which serves as the inspiration for the name ``agent environment cycle''.

\section{API Design}

\subsection{Basic API}
\label{sec:basic-api}
The PettingZoo API is shown in \autoref{fig:pettingzoo_example}, and the strong similarities to the Gym API (\autoref{fig:gym_example}) should be obvious --- each agent provides an \texttt{action} to a \texttt{step} function and receives \texttt{observation, reward, done, info} as the return values. The observation and state spaces also use the the exact same space objects as Gym. 
The \texttt{render} and \texttt{close} methods also function identically to Gym's, showing a current visual frame representing the environment to the screen whenever called. The \texttt{reset} method similarly has identical function to Gym --- it resets the environment to a starting configuration after being played through. PettingZoo really only has two deviations from the regular Gym API --- the \texttt{last} and \texttt{agent\_iter} methods and the corresponding iteration logic.

\begin{figure}[ht]
\begin{center}
\begin{BVerbatim}[commandchars=\\\{\},fontsize=\scriptsize]
\PY{k+kn}{from} \PY{n+nn}{pettingzoo}\PY{n+nn}{.}\PY{n+nn}{butterfly} \PY{k+kn}{import} \PY{n}{pistonball\PYZus{}v0}
\PY{n}{env} \PY{o}{=} \PY{n}{pistonball\PYZus{}v0}\PY{o}{.}\PY{n}{env}\PY{p}{(}\PY{p}{)}
\PY{n}{env}\PY{o}{.}\PY{n}{reset}\PY{p}{(}\PY{p}{)}
\PY{k}{for} \PY{n}{agent} \PY{o+ow}{in} \PY{n}{env}\PY{o}{.}\PY{n}{agent\PYZus{}iter}\PY{p}{(}\PY{l+m+mi}{1000}\PY{p}{)}\PY{p}{:}
    \PY{n}{env}\PY{o}{.}\PY{n}{render}\PY{p}{(}\PY{p}{)}
    \PY{n}{observation}\PY{p}{,} \PY{n}{reward}\PY{p}{,} \PY{n}{done}\PY{p}{,} \PY{n}{info} \PY{o}{=} \PY{n}{env}\PY{o}{.}\PY{n}{last}\PY{p}{(}\PY{p}{)}
    \PY{n}{action} \PY{o}{=} \PY{n}{policy}\PY{p}{(}\PY{n}{observation, agent}\PY{p}{)}
    \PY{n}{env}\PY{o}{.}\PY{n}{step}\PY{p}{(}\PY{n}{action}\PY{p}{)}
\PY{n}{env}\PY{o}{.}\PY{n}{close}\PY{p}{(}\PY{p}{)}
\end{BVerbatim}
\caption{An example of the basic usage of Pettingzoo}
\label{fig:pettingzoo_example}
\end{center}
\end{figure}

\subsection{The \texttt{agent\_iter} Method}
The \texttt{agent\_iter} method is a generator method of an environment that returns the next agent that the environment will be acting upon. Because the environment is providing the next agent to act, this cleanly abstracts away any issues surrounding changing agent orders, agent generation, and agent death. This generation also parallels the functionality of the next agent function from the AEC games model. This method, combined with one agent acting at once, allows for the support of every conceivable variation of the set of agents changing.

\subsection{The \texttt{last} Method}
An odd aspect of multi-agent environments is that from the perspective of one agent, the other agents are part of the environment. Whereas in the single agent case the observation and rewards can be given immediately, in the multi-agent case an agent has to wait for all other agents to act before it's \texttt{observation}, \texttt{reward}, \texttt{done} and \texttt{info} can be fully determined. For this reason, these values are given by the \texttt{last} method, and they can then be passed into a policy to choose an action. Less robust implementations would not allow for features like changing agent orders (like the reverse card in Uno).

\subsection{Additional API Features}
\label{sec:additional_api}

The \texttt{agents} attribute is a list of all agents in the environment, as strings. The \texttt{rewards}, \texttt{dones}, \texttt{infos} attributes are agent-keyed dictionaries for each attribute (note that the rewards are the instantaneous ones resulting from the most recent action). These allow access to agent properties at all points on a trajectory, regardless of which is selected. The \texttt{action\_space(agent)} and \texttt{observation\_space(agent)} functions return the static action and observation spaces respectively for the agent given as an argument.  The \texttt{observe(agent)} method provides the observation for a single agent by passing its name as an argument, which can be useful if you need to observe an agent in an unusual context. The \texttt{state} method is an optional method returns the global state of an environment, as is required for centralized critic methods. The \texttt{agent\_selection} method returns the agent that can currently be acted upon per \texttt{agent\_iter}.

The motivation for allowing access to all these lower level pieces of information is to let researchers to attempt novel, unusual experiments. The space of multi-agent RL has not yet been comprehensively explored, and there are many perfectly plausible reasons you might want access to other agents rewards, observations, and so on. For an API to be universal in an emerging field, it inherently has to allow access to all the information researchers could plausibly want. For this reason we allow access to a fairly straightforward set of lower level attributes and methods in addition to the standard higher level API. As we outline in \autoref{sec:env_creation}, we've structured PettingZoo in a way such that including these low-level features doesn't introduce engineering overhead in creating environments, as discussed further in the documentation website.

To handle environments where different agents can be present on each reset of an environment, PettingZoo has an optional \texttt{possible\_agents} attribute which lists all the agents that might exist in an environment at any point. Environments which generate arbitrary numbers or types of agents will not define a \texttt{possible\_agents} list, requiring the user to check for new agents being instantiated as the environment runs. After resetting the environment, the \texttt{agents} attribute becomes accessible and lists all agents that are currently active. For similar reasons, \texttt{num\_agents}, \texttt{rewards}, \texttt{dones}, \texttt{infos}, and \texttt{agent\_selection} are not available until after a reset.

To handle cases where environments need to have environment agents as per the formal AEC Games model, the standard is to put it into the \texttt{agents} with the name \texttt{env} and have it take \texttt{None} as it's action. We do not require this for all environments by default as it's rarely used and makes the API more cumbersome, but this is an important feature for certain edge cases in research. This connects to the formal model in that, when this feature is not used, the environment actor from the formal model and the agent actor that acted before it are merged together.

\subsection{Environment Creation and the Parallel API}
\label{sec:env_creation}
PettingZoo environments actually only expose the  \texttt{reset}, \texttt{seed}, \texttt{step}, \texttt{observe}, \texttt{render}, and \texttt{close} base methods and the \texttt{agents}, \texttt{rewards}, \texttt{dones}, \texttt{infos}, \texttt{state} and \texttt{agent\_iter} base attributes. These are then wrapped to add the \texttt{last} method. Only having environments implement primitive methods makes creating new environments simpler, and reduces code duplication. This has the useful side effect of allowing all PettingZoo environments to be easily changed to an alternative API by simply writing a new wrapper. We've actually already done this for the default environments and added an additional ``parallel API'' to them that's almost identical to the RLlib POSG-based API via a wrapper. We added this secondary API because in environments with very large numbers of agents, this can improve runtime by reducing the number of Python function calls.

\section{Default Environments}
Similar to Gym's default environments, PettingZoo includes 63 environments. Half of the included environment classes (MPE, MAgent, and SISL), despite their popularity, existed as unmaintained ``research grade'' code, have not been available for installation via pip, and have required large amounts of maintenance to run at all before our cleanup and maintainership. We additionally included multiplayer Atari games from \citet{terry2020arcade}, Butterfly environments which are original and of our own creation, and popular classic board and card game environments. All default environments included are surveyed in depth in \autoref{app:default_envs}.

\section{Adoption}
In it's relatively short lifespan, PettingZoo has already achieved a meaningful amount of adoption. 
It is supported by the following learning libraries: The Autonomous Learning Library \citep{nota2020autonomous}, AI-Traineree \citep{AI-Traineree}, PyMARL (ongoing) \citep{smac2019}, RLlib \citep{liang2018rllib}, Stable Baselines 2 \citep{stable-baselines} and Stable Baselines 3 \citep{stable-baselines3}, similar libraries such as CleanRL \citep{cleanrl} (through SuperSuit \citep{SuperSuit}), and Tianshou (ongoing) \citep{tianshou}. 
Perhaps more significantly than any of this, PettingZoo is already being used to teach in both graduate and undergraduate reinforcement learning classes all over the world.

\section{Conclusion}

This paper introduces PettingZoo, a Python library of many diverse multi-agent reinforcement learning environments under one simple API, akin to a multi-agent version of OpenAI's Gym library, and introduces the agent environment cycle game model of multi-agent games. 

Given the importance of multi-agent reinforcement learning, we believe that PettingZoo is capable of democratizing the field similar to what Gym previously did for single agent reinforcement learning, making it accessible to university scale research and to non-experts. As evidenced by it's early adoption into numerous MARL libraries and courses, PettingZoo is moving in the direction of accomplishing this goal.

We're aware of one notable limitation of the PettingZoo API. Games with significantly more than 10,000 agents (or potential agents) will have meaningful performance issues because you have to step each agent at once. Efficiently updating environments like this, and inferencing with the associated policies, requires true parallel support which almost certainly should be done in a language other than Python. Because of this, we view this as a practically acceptable limitation.

We see three directions for future work. The first is additions of more interesting environments under our API (possibly from the community, as has happened with Gym). The second direction we envision is a service to allow different researchers' agents to play against each other in competitive games, leveraging the standardized API and environment set. Finally, we envision the development of procedurally generated multi-agent environments to test how well methods generalize, akin to the Gym procgen environments \citep{cobbe2019leveraging}.

\section*{Acknowledgements}

J.K. Terry was supported during part of this work by the QinetiQ Fundamental Machine Learning Fellowship. Thank you to Kyle Sang for their contributions to the documentation website. Thank you Rohan Potdar and Sang Hyun Son for their contributions to the Butterfly benchmarks. Thank you to Deepthi Raghunandan and Christian Clauss for their contributions to testing and continuous integration. Thank you to the PettingZoo community for the numerous bug reports and contributions to the package, especially Ross Allen and their group.

\bibliographystyle{plainnat}
\bibliography{main}

\newpage

\appendix

\section{Additional Case Study Information} \label{app:casestudies}

\subsection{Race Conditions in Sequential Social Dilemma Games}
\label{sec:appendix_bug_ssd}

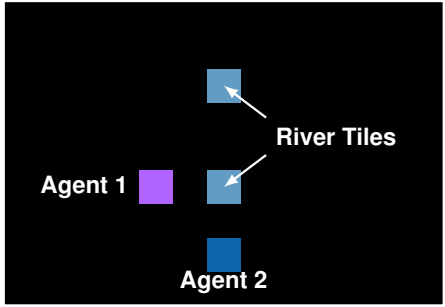
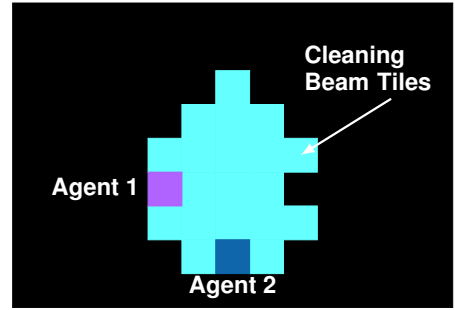
\begin{figure}[!tbp]
\centering
\begin{subfigure}{0.45\linewidth}
\begin{tikzpicture}[x=\sz,y=\sz, every node/.style={white,minimum size=\sz}, every path/.style={white,line width=0.3mm,font={\bfseries\sffamily\footnotesize}},>=latex]
\draw[fill=black] (-.5,-.5) rectangle (12.5,8.5);

\node[label={180:Agent 1},fill=agent1,rectangle] (a1) at (4,3) {};
\node[fill=agent2,rectangle,label={[label distance=-.15cm]-90:Agent 2}] (a2) at (6,1) {};

\node[fill=river] (r0) at (6,3) {};
\node[fill=river] (r1) at (6,6) {};

\node[anchor=west] (rtt) at (7.25,4.5) {River Tiles};
\draw[->] (rtt.south west) -- (r0.center);
\draw[->] (rtt.north west) -- (r1.center);
\end{tikzpicture}




%
\caption{The initial setup with two agents and two river tiles. When the river tiles become dirty, they are shown as a brownish color instead.}%
\end{subfigure}%
\hfill
\begin{subfigure}{0.45\linewidth}
\begin{tikzpicture}[x=\sz,y=\sz, every node/.style={white,minimum size=\sz}, every path/.style={ultra thin}, >=latex, font={\bfseries\sffamily\footnotesize}]
\draw[fill=black] (-.5,-.5) rectangle (12.5,8.5); 

\node[draw=cleaningbeam,fill=cleaningbeam] () at (5,1) {};
\node[draw=cleaningbeam,fill=cleaningbeam] () at (7,1) {};
\node[draw=cleaningbeam,fill=cleaningbeam] () at (4,2) {};
\node[draw=cleaningbeam,fill=cleaningbeam] () at (5,2) {};
\node[draw=cleaningbeam,fill=cleaningbeam] () at (6,2) {};
\node[draw=cleaningbeam,fill=cleaningbeam] () at (7,2) {};
\node[draw=cleaningbeam,fill=cleaningbeam] () at (8,2) {};
\node[draw=cleaningbeam,fill=cleaningbeam] () at (5,3) {};
\node[draw=cleaningbeam,fill=cleaningbeam] () at (6,3) {};
\node[draw=cleaningbeam,fill=cleaningbeam] () at (7,3) {};
\node[draw=cleaningbeam,fill=cleaningbeam] () at (4,4) {};
\node[draw=cleaningbeam,fill=cleaningbeam] () at (5,4) {};
\node[draw=cleaningbeam,fill=cleaningbeam] () at (6,4) {};
\node[draw=cleaningbeam,fill=cleaningbeam] () at (7,4) {};
\node[draw=cleaningbeam,fill=cleaningbeam] (cb) at (8,4) {};
\node[draw=cleaningbeam,fill=cleaningbeam] () at (5,5) {};
\node[draw=cleaningbeam,fill=cleaningbeam] () at (6,5) {};
\node[draw=cleaningbeam,fill=cleaningbeam] () at (7,5) {};
\node[draw=cleaningbeam,fill=cleaningbeam] () at (6,6) {};

\node[anchor=north east, text width=2.3cm] (cbtt) at (13.5,7.5) {Cleaning Beam Tiles};

\node[label={180:Agent 1},draw=agent1,fill=agent1,rectangle] (a1) at (4,3) {};
\node[draw=agent2,fill=agent2,rectangle,label={[label distance=-.09cm]-90:Agent 2}] (a2) at (6,1) {};

\draw[white,->,line width=0.3mm] (cbtt.south) -- (cb.center);

\end{tikzpicture}%
\caption{The result of both agents perform the ``clean'' action. Both river tiles can be are cleaned since Agent 1's action is resolved first.}%
\end{subfigure}%
\caption{Cleanup, a Sequential Social Dilemma Game from~\citet{SSDOpenSource}.}
\label{fig:ssd1}
\end{figure}

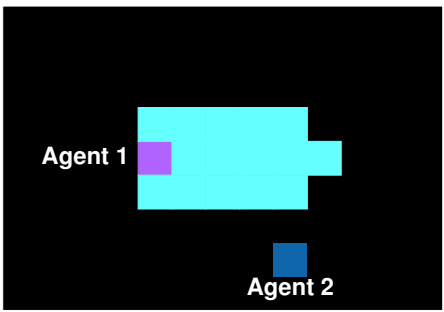
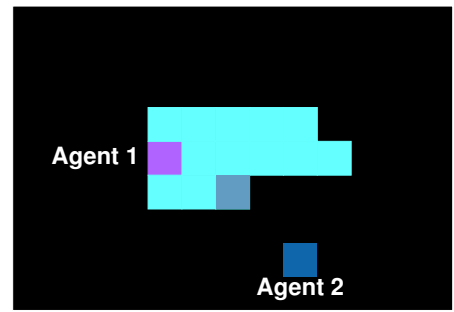
\begin{figure}[!tbp]
\centering
\begin{subfigure}[b]{0.45\linewidth}
    \begin{tikzpicture}[x=\sz,y=\sz, every node/.style={white, minimum size=\sz}, every path/.style={white, ultra thin, font={\bfseries\sffamily\footnotesize}}, >=latex]
\draw[fill=black] (-.5,-.5) rectangle (12.5,8.5);

\node[label={180:Agent 1},draw=agent1,fill=agent1,rectangle] (a1) at (4,4) {};

\node[draw=cleaningbeam,fill=cleaningbeam] () at (5,4) {};
\node[draw=cleaningbeam,fill=cleaningbeam] () at (6,4) {};
\node[draw=cleaningbeam,fill=cleaningbeam] () at (7,4) {};
\node[draw=cleaningbeam,fill=cleaningbeam] () at (8,4) {};
\node[draw=cleaningbeam,fill=cleaningbeam] () at (9,4) {};

\node[draw=cleaningbeam,fill=cleaningbeam] () at (4,3) {};
\node[draw=cleaningbeam,fill=cleaningbeam] () at (5,3) {};
\node[draw=cleaningbeam,fill=cleaningbeam] () at (6,3) {};
\node[draw=cleaningbeam,fill=cleaningbeam] () at (7,3) {};
\node[draw=cleaningbeam,fill=cleaningbeam] () at (8,3) {};

\node[draw=cleaningbeam,fill=cleaningbeam] () at (4,5) {};
\node[draw=cleaningbeam,fill=cleaningbeam] () at (5,5) {};
\node[draw=cleaningbeam,fill=cleaningbeam] () at (6,5) {};
\node[draw=cleaningbeam,fill=cleaningbeam] () at (7,5) {};
\node[draw=cleaningbeam,fill=cleaningbeam] () at (8,5) {};

\node[fill=agent2,rectangle,label={[label distance=-.1cm]-90:Agent 2}] (a2) at (8,1) {};

\end{tikzpicture}
    \caption{If there are no dirty river tiles in the path of the cleaning beams, the beams will extend to the full length of five tiles.}
    \label{fig:ssdbeama}
\end{subfigure}
\hfill
\begin{subfigure}[b]{0.45\linewidth}
    \begin{tikzpicture}[x=\sz,y=\sz, every node/.style={white, minimum size=\sz}, every path/.style={white, ultra thin, font={\bfseries\sffamily\footnotesize}}, >=latex]
\draw[fill=black] (-.5,-.5) rectangle (12.5,8.5);

\node[label={180:Agent 1},draw=agent1,fill=agent1,rectangle] (a1) at (4,4) {};

\node[draw=cleaningbeam,fill=cleaningbeam] () at (5,4) {};
\node[draw=cleaningbeam,fill=cleaningbeam] () at (6,4) {};
\node[draw=cleaningbeam,fill=cleaningbeam] () at (7,4) {};
\node[draw=cleaningbeam,fill=cleaningbeam] () at (8,4) {};
\node[draw=cleaningbeam,fill=cleaningbeam] () at (9,4) {};

\node[draw=cleaningbeam,fill=cleaningbeam] () at (4,3) {};
\node[draw=cleaningbeam,fill=cleaningbeam] () at (5,3) {};
\node[draw=cleaningbeam,fill=cleaningbeam] () at (6,3) {};

\node[draw=cleaningbeam,fill=cleaningbeam] () at (4,5) {};
\node[draw=cleaningbeam,fill=cleaningbeam] () at (5,5) {};
\node[draw=cleaningbeam,fill=cleaningbeam] () at (6,5) {};
\node[draw=cleaningbeam,fill=cleaningbeam] () at (7,5) {};
\node[draw=cleaningbeam,fill=cleaningbeam] () at (8,5) {};

\node[fill=river] () at (6,3) {};

\node[fill=agent2,rectangle,label={[label distance=-.1cm]-90:Agent 2}] (a2) at (8,1) {};

\end{tikzpicture}
    \caption{If there is a dirty river tile in the path of a beam, the beam will stop at the tile, changing it to a ``clean'' river tile.}
\end{subfigure}
\caption{An example of Agent 1 using the ``clean'' action while facing East. The beams extend to a length of up to five tiles. The ``main'' beam extends directly in front of the agent, while two auxiliary beams start at the tiles directly next to the agent (one to the left and one to the right) and also extend up to five tiles. A beam stops when it hits a dirty river tile. }
\label{fig:ssdbeam}
\end{figure}

The Sequential Social Dilemma Games, introduced in~\cite{leibo2017multi}, are a kind of MARL environment where good short-term strategies for single agents lead to bad long-term results for all of the agents. New SSD environments, including the \emph{Cleanup} environment, were introduced in~\citet{hughes2018inequity}. All of these have open source implementations in ~\citep{SSDOpenSource}. The states of these games are represented by a grid of tiles, where each tile represents either an agent or a piece of the environment. In the \emph{Cleanup} environment, the environment tiles can be empty tiles, river tiles, and apple tiles. Collecting apple tiles results in a reward for the agent and the agents must clean the river tiles with a ``cleaning beam'' for apple tiles to spawn. The cleaning beam extends in front of agents, one tile at a time, until it hits a dirty river tile (``waste'') or extends to its maximum length of 5 tiles. Additionally, two more beams extend in front of the agent---one starting in the tile directly to the agent's left, and one from the tile on the right---until each hits a ``waste'' tile or reaches a length of 5 tiles. The cleaning beam is shown in \autoref{fig:ssdbeama}. Note that while beams stop at ``waste'' tiles, they will continue to extend past clean river tiles.

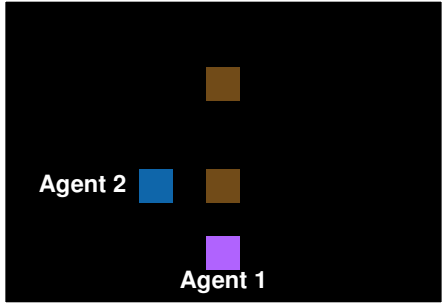
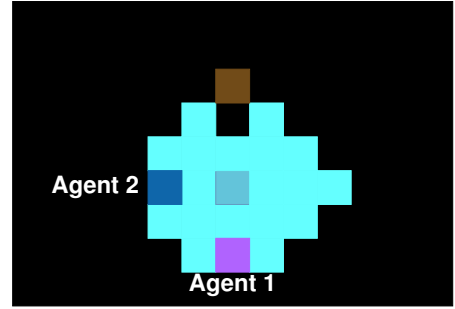
\begin{figure}[!tbp]
\centering
\begin{subfigure}[b]{0.45\linewidth}
    \begin{tikzpicture}[x=\sz,y=\sz, every node/.style={white, minimum size=\sz}, every path/.style={white, line width=0.5mm}, >=latex, font={\bfseries\sffamily\footnotesize}]
\draw[fill=black] (-.5,-.5) rectangle (12.5,8.5);

\node[label={180:Agent 2},fill=agent2,rectangle] (a2) at (4,3) {};
\node[fill=agent1,rectangle,label={[label distance=-.15cm]-90:Agent 1}] (a1) at (6,1) {};

\node[fill=waste] (r0) at (6,3) {};
\node[fill=waste] (r1) at (6,6) {};
\end{tikzpicture}
    \caption{The same setup as in \autoref{fig:ssd1}, but with the agent labels reversed.}
\end{subfigure}
\hfill
\begin{subfigure}[b]{0.45\linewidth}
        \begin{tikzpicture}[x=\sz,y=\sz, every node/.style={white, minimum size=\sz}, every path/.style={ultra thin}, >=latex, font={\bfseries\sffamily\footnotesize}]
\draw[fill=black] (-.5,-.5) rectangle (12.5,8.5); 

\node[draw=cleaningbeam,fill=cleaningbeam] () at (5,1) {};
\node[draw=cleaningbeam,fill=cleaningbeam] () at (7,1) {};
\node[draw=cleaningbeam,fill=cleaningbeam] () at (4,2) {};
\node[draw=cleaningbeam,fill=cleaningbeam] () at (5,2) {};
\node[draw=cleaningbeam,fill=cleaningbeam] () at (6,2) {};
\node[draw=cleaningbeam,fill=cleaningbeam] () at (7,2) {};
\node[draw=cleaningbeam,fill=cleaningbeam] () at (8,2) {};
\node[draw=cleaningbeam,fill=cleaningbeam] () at (5,3) {};
\node[draw=river,fill=river!60!cleaningbeam] () at (6,3) {};
\node[draw=cleaningbeam,fill=cleaningbeam] () at (7,3) {};
\node[draw=cleaningbeam,fill=cleaningbeam] () at (8,3) {};
\node[draw=cleaningbeam,fill=cleaningbeam] () at (9,3) {};
\node[draw=cleaningbeam,fill=cleaningbeam] () at (4,4) {};
\node[draw=cleaningbeam,fill=cleaningbeam] () at (5,4) {};
\node[draw=cleaningbeam,fill=cleaningbeam] () at (6,4) {};
\node[draw=cleaningbeam,fill=cleaningbeam] () at (7,4) {};
\node[draw=cleaningbeam,fill=cleaningbeam] (cb) at (8,4) {};
\node[draw=cleaningbeam,fill=cleaningbeam] () at (5,5) {};
\node[draw=cleaningbeam,fill=cleaningbeam] () at (7,5) {};

\node[draw=waste,fill=waste] () at (6,6) {};

\node[label={180:Agent 2},draw=agent2,fill=agent2,rectangle] (a2) at (4,3) {};
\node[draw=agent1,fill=agent1,rectangle,label={[label distance=-.09cm]-90:Agent 1}] (a1) at (6,1) {};
\end{tikzpicture}
    \caption{The result of both agents performing the ``clean'' action, with this agent assignment.}
\end{subfigure}
\caption{The impact of switching the internal agent order on how the environment evolves. When both agents clean, agent 1's action is resolved first, and the main beam stops when it hits the near dirty river tile, so the far river tile is not cleaned. In \autoref{fig:ssd1}, Agent 2's beam was able to reach the far beam because Agent 1's beam cleaned the near tile first.}
\label{fig:ssd2}
\end{figure}

The agents act sequentially in the same order every turn, including the firing of their beams. In the case of two agents trying to occupy the same space, one is chosen randomly, however the tie breaking with regards to the beams is biased, due to a bug. Consider the setup in \autoref{fig:ssd1} where each agent chooses the ``clean'' action for the next step. This results in Agent 1 firing their cleaning beam first, clearing the close river tile. Next, Agent 2 fires their cleaning beam and they are able to clean the far river tile because the close tile has already been cleared by Agent 1. However, if we keep the same placement and actions but switch the labels of the agents, we get a different result, seen in \autoref{fig:ssd2}. Now, Agent 1 fires first and hits the close river tile and can no longer reach the far river tile. In situations like these, the observation the second agent's policy is using to act on is going to be inherently wrong, and if it had the true environment state before acting it would very likely wish to make a different choice.

This is a serious class of bug that's very easy to introduce when using parallel action-based APIs, while using AEC games-based APIs prevents the class entirely. In this specific instance, the bug had gone unnoticed for years.

\subsection{Reward Defects in Pursuit}

\label{sec:appendix_bug_pursuit}

We validated the impact of reward pruning experimentally by training parameter shared Ape-X DQN \citep{horgan2018distributed} (the best performing model on pursuit~\citep{TerryParameterSharing}) four times using RLLib \citep{liang2017rllib} with and without reward pruning, achieving better results with reward pruning every time and 22.03\% more total reward on average \autoref{fig:pursuit_ppo}, while PPO \citep{schulman2017proximal} learned 16.12\% more reward on average with this \autoref{fig:pporewardpruningresults}. Saved training logs and all code needed to reproduce the experiments and plots is available in the supplemental materials.

\begin{figure}[ht]
\centering
\begin{subfigure}{0.98\linewidth}
\centering
\scalebox{0.65}{
\input{figures/pruned_pursuit_adqn.pgf}
}
\caption{Learning on the \emph{pursuit} environment with and without pruned rewards, using parameter sharing based on Ape-X DQN. This shows an average of an 22.03\% improvement by using this method.} 
\label{fig:pursuit_ppo}
\end{subfigure}

%
\begin{subfigure}{0.98\linewidth}
\centering
\scalebox{0.65}{
\input{figures/pruned_pursuit_ppo.pgf}
}
\caption{Learning on the \emph{pursuit} environment with and without reward pruning, using parameter sharing based on PPO. Reward pruning increased the total reward by 16.12\% on average.}
\label{fig:pporewardpruningresults}
\end{subfigure}
\label{fig:something}
\end{figure}

\section{Default Environments}
\label{app:default_envs}
This section surveys all the environments that are included in PettingZoo by default.

\begin{figure*}[ht]
\vskip 0.2in
\begin{center}
\begin{subfigure}{\textwidth}
\begin{subfigure}{0.49\textwidth}
\centering
\includegraphics[scale=0.5]{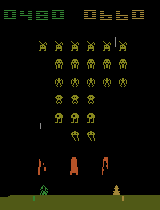}
\caption{Atari: Space Invaders}
\label{figure:atari}
\end{subfigure}
\begin{subfigure}{0.5\textwidth}
\centering
\includegraphics[scale=0.15]{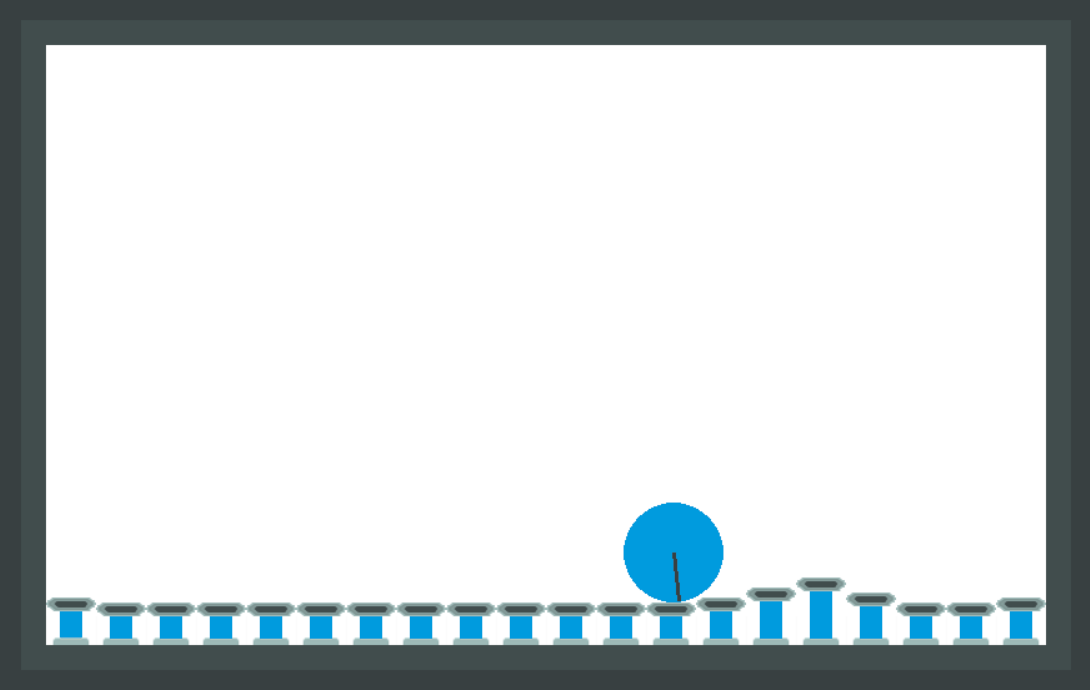}
\caption{Butterfly: Pistonball}
\label{figure:butterfly}
\end{subfigure}
\end{subfigure}

\begin{subfigure}{\textwidth}
\begin{subfigure}{0.49\textwidth}
\centering
\includegraphics[scale=0.4]{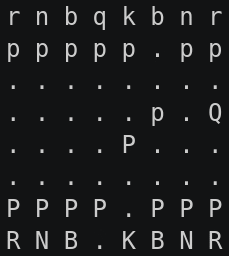}
\caption{Classic: Chess}
\label{figure:classic}
\end{subfigure}
\begin{subfigure}{0.5\textwidth}
\centering
\includegraphics[scale=0.3]{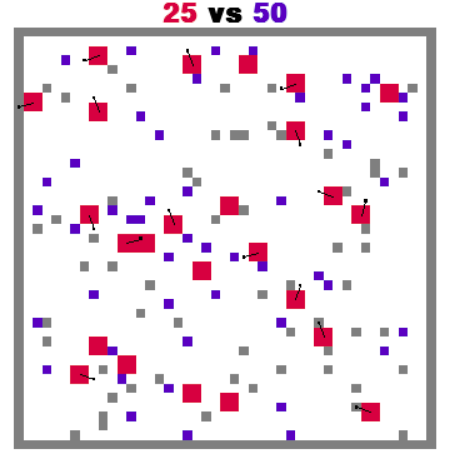}
\caption{MAgent: Adversarial Pursuit}
\label{figure:magent}
\end{subfigure}
\end{subfigure}

\begin{subfigure}{\textwidth}
\begin{subfigure}{0.48\textwidth}
\centering
\includegraphics[scale=0.18]{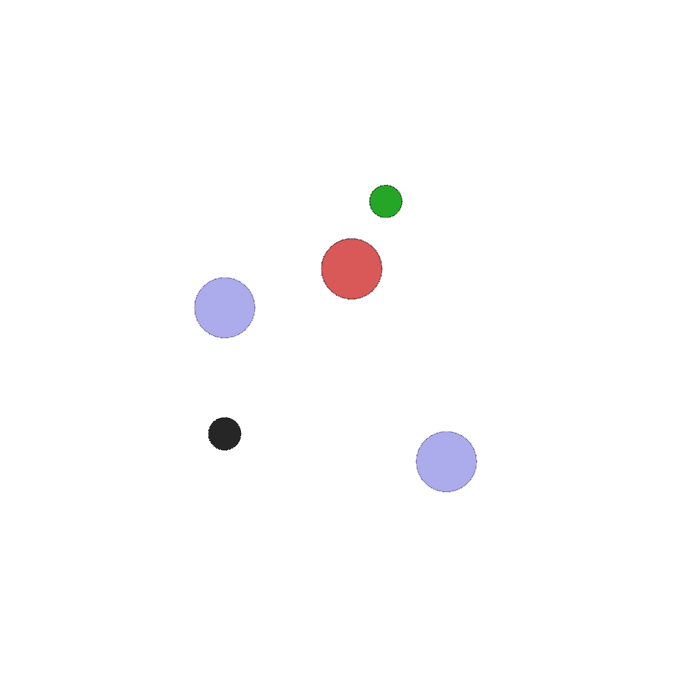}
\caption{MPE: Simple Adversary}
\label{figure:mpe}
\end{subfigure}
\begin{subfigure}{0.5\textwidth}
\centering
\includegraphics[scale=0.25]{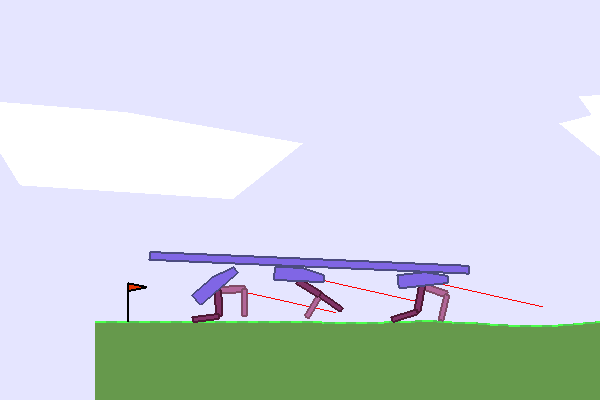}
\caption{SISL: Multiwalker}
\label{figure:sisl}
\end{subfigure}
\end{subfigure}
\caption{Example environments from each class.}
\end{center}
\end{figure*}

\textbf{Atari}

Atari games represent the single most popular and iconic class of benchmarks in reinforcement learning. Recently, a multi-agent fork of the Arcade Learning Environment was created that allows programmatic control and reward collection of Atari's iconic multi-player games \citep{terry2020arcade}. 
As in the single player Atari environments, the observation is the rendered frame of the game, which is shared between all agents, so there is no partial observability. 
Most of these games have competitive or mixed reward structures, making them suitable for general study of adversarial and mixed reinforcement learning. In particular, \citet{terry2020arcade} categorizes the games into 7 different types: 1v1 tournament games, mixed sum survival games (\emph{Space Invaders}, shown in Figure \ref{figure:atari}. is an example of this), competitive racing games, long term strategy games, 2v2 tournament games, a four-player free-for-all game and a cooperative game. 

\textbf{Butterfly}

Of all the default environments included, the majority of them are competitive. We wanted to supplement this with a set of interesting graphical cooperative environments. \emph{Pistonball}, depicted in Figure \ref{figure:butterfly}, is an environment where pistons need to coordinate to move a ball to the left, while only being able to observe a local part of the screen. It requires learning nontrivial emergent behavior and indirect communication to perform well. \emph{Knights Archers Zombies} is a game in which agents work together to defeat approaching zombies before they can reach the agents. It is designed to be a fast paced, graphically interesting combat game with partial observability and heterogeneous agents, where achieving good performance requires extraordinarily high levels of agent coordination. In \emph{Cooperative pong} two dissimilar paddles work together to keep a ball in play as long as possible. It was intended to be a be very simple cooperative continuous control-type task, with heterogeneous agents. \emph{Prison} was designed to be the simplest possible game in MARL, and to be used as a debugging tool. In this environment, no agent has any interaction with the others, and each agent simply receives a reward of 1 when it paces from one end of its prison cell to the other. \emph{Prospector} was created to be a very challenging game for conventional methods---it has two classes of agents with different goals, action spaces, and observation spaces (something many current cooperative MARL algorithms struggle with), and has very sparse rewards (something all RL algorithms struggle with). It is intended to be a very difficult benchmark for MARL, in the same vein as Montezuma's Revenge.

\textbf{Classic}

Classical board and card games have long been some of the most popular environments in reinforcement learning \citep{gerald1995temporal,silver2016mastering,bard2019hanabi}. We include all of the standard multiplayer games in RLCard \citep{zha2019rlcard}: \emph{Dou Dizhu}, \emph{Gin Rummy}, \emph{Leduc Hold’em}, \emph{Limit Texas Hold'em}, \emph{Mahjong}, \emph{No-limit Texas Hold'em}, and \emph{Uno}. We additionally include all AlphaZero games, using the same observation and action spaces---\emph{Chess} and \emph{Go}. We finally included \emph{Backgammon}, \emph{Connect Four}, \emph{Checkers}, \emph{Rock Paper Scissors}, \emph{Rock Paper Scissors Lizard Spock}, and \emph{Tic Tac Toe} to add a diverse set of simple, popular games to allow for more robust benchmarking of RL methods.

\textbf{MAgent}

The MAgent library, from \citet{zheng2017magent} was introduced as a configurable and scalable environment that could support thousands of interactive agents. These environments have mostly been studied as a setting for emergent behavior \citep{Pokle2018AnalysisOE}, heterogeneous agents \citep{Subramanian2020MultiTM}, and efficient learning methods with many agents \citep{Chen2019FactorizedQF}. We include a number of preset configurations, for example the \emph{Adversarial Pursuit} environment shown in Figure \ref{figure:magent}. We make a few changes to the preset configurations used in the original MAgent paper. The global "minimap" observations in the battle environment are turned off by default, requiring implicit communication between the agents for complex emergent behavior to occur. The rewards in \emph{Gather} and \emph{Tiger-Deer} are also slightly changed to prevent emergent behavior from being a direct result of the reward structure. 

\textbf{MPE}

The Multi-Agent Particle Environments (MPE) were introduced as part of \citet{mordatch2017emergence} and first released as part of \citet{lowe2017multi}. These are 9 communication oriented environments where particle agents can (sometimes) move, communicate, see each other, push each other around, and interact with fixed landmarks. Environments are cooperative, competitive, or require team play. They have been popular in research for general MARL methods \citet{lowe2017multi}, emergent communication \citep{mordatch2017emergence}, team play \citep{Palmer2020IndependentLA}, and much more. As part of their inclusion in PettingZoo, we converted the action spaces to a discrete space which is the Cartesian product of the movement and communication action possibilities. We also added comprehensive documentation, parameterized any local reward shaping (with the default setting being the same as in \citet{lowe2017multi}), and made a single render window which captures all the activities of all agents (including communication), making it easier to visualize.

\textbf{SISL}

We finally included the three cooperative environments introduced in \citet{gupta2017cooperative}: \emph{Pursuit}, \emph{Waterworld}, and \emph{Multiwalker}. \emph{Pursuit} is a standard pursuit-evasion game \citet{vidal2002probabilistic} where pursuers are controlled in a randomly generated map. Pursuer agents are rewarded for capturing randomly generated evaders by surrounding them on all sides. \emph{Waterworld} is a continuous control game where the pursuing agents cooperatively hunt down food targets while trying to avoid poison targets. \emph{Multiwalker} (Figure \ref{figure:sisl}) is a more challenging continuous control task that is based on Gym's \emph{BipedalWalker} environment. In \emph{Multiwalker}, a package is placed on three independently controlled robot legs. Each robot is given a small positive reward for every unit of forward horizontal movement of the package, while they receive a large penalty for dropping the package.

\subsection{Butterfly Baselines}
\label{section:baselines}

Whne environments are introduced to the literature, it is customary for them to include baselines to provide a general sense of the difficulty of the environment and to provide something to compare against. We do this here for the Butterfly environments that this library introduces for the first time; similar baselines exist in the papers introducing all other environments. For our baseline learning method we used used fully parameter shared PPO \citep{schulman2017proximal} from Stable-Baselines3 (SB3) \citep{stable-baselines3}. We use the SuperSuit wrapper library \citep{SuperSuit2020} for preprocessing similar to that in \cite{mnih2015human}, convert the observations to grayscale, resize them to 96x96 images, and use frame-stacking to combine the last four observations. Furthermore, for cooperative\_pong\_v3 and knights\_archers\_zombies\_v7, we invert the color of alternating agent's observations by subtracting it from the maximum observable value to improve learning and differentiate which agent type an observation came from for the parameter shared neural network, per \citet{terry2020revisiting}. On the prospector\_v4 environment, we add an extra channel to the observations which is set to the maximum possible value if the agent belongs to the opposite agent type, else zero. Both these modifications allow us to use parameter-shared PPO across non-homogeneous agents. On prospector\_v4 we also pad observation and agent spaces as described in \citet{terry2020revisiting} to allow for learning with a single fully parameter shared neural network.

\begin{figure}[!hbt]
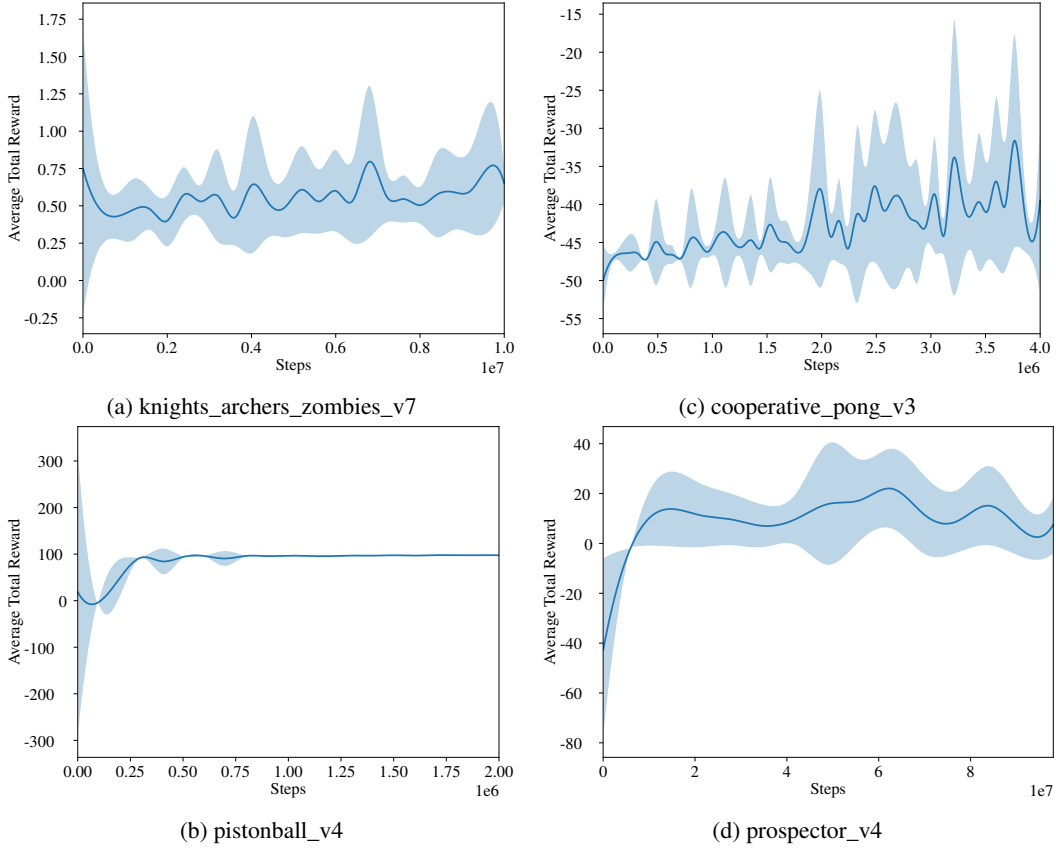

\centering
\begin{minipage}{0.49\textwidth}
\begin{subfigure}{\linewidth}
  \centering
  \resizebox{\linewidth}{!}{\input{figures/PPO_knights_archers_zombies_v7.pgf}}
  \caption{knights\_archers\_zombies\_v7}
\end{subfigure}
\begin{subfigure}{\linewidth}
  \centering
  \resizebox{\linewidth}{!}{\input{figures/PPO_pistonball_v4.pgf}}
  \caption{pistonball\_v4}
\end{subfigure}
\end{minipage}
\hfill
\begin{minipage}{0.49\textwidth}
\begin{subfigure}{\linewidth}
  \centering
  \resizebox{\linewidth}{!}{\input{figures/PPO_cooperative_pong_v3.pgf}}
  \caption{cooperative\_pong\_v3}
\end{subfigure}
\begin{subfigure}{\linewidth}
  \centering
  \resizebox{\linewidth}{!}{\input{figures/PPO_prospector_v4.pgf}}
  \caption{prospector\_v4}
\end{subfigure}
\end{minipage}
\caption{Total reward when learning on each Butterfly environment via parameter-shared PPO.}
\label{fig:butterfly_results}
\end{figure}

After tuning hyperparameters with RL Baselines3 Zoo \citep{rl-zoo3}, our baselines learns an optimal policy in the Pistonball environment and  Cooperative Pong environments and learns reasonably in the Knights Archers Zombies and Prospector environments without achieving optimal policies. Plots showing results of 10 training runs of the best hyperparameters  are shown in \autoref{fig:butterfly_results}. All code and hyperparameters for these runs is available at \url{https://github.com/jkterry1/Butterfly-Baselines}.

\section{Formal Definitions}
\label{sec:formaldefs}

\subsection{Partially Observable Stochastic Games}
\label{sec:posgdef}

  The formal definition of a POSG is shown in \autoref{def:posg}. This definition can be viewed as the typical Stochastic Games model~\citep{shapley1953stochasticgames} with the addition of POMDP-style partial observability.

\begin{definition}
\label{def:posg}
  A \emph{Partially-Observable Sto\-chas\-tic Game} (POSG) is a tuple
  $\langle \states, s_0, \aN, (\actions_{i})_{i\in\agentset}
  ,\transition, (\reward_{i})_{i\in\agentset},
  ,
  (\observations_{i})_{i\in\agentset},
  ,
  (\obsfunc_{i})_{i\in\agentset}
  \rangle$, where:
  \begin{itemize}
  \item $\states$ is the set of possible \emph{states}.
  \item $s_0$ is the \emph{initial state}.
  \item $\aN$ is the \emph{number of agents}. The \emph{set of
      agents} is $\agentset$.
  \item $\actions_{i}$ is the set of possible \emph{actions} for agent
    $i$.
  \item
    $\transition\colon {{\states \times 
        {\prod_{i\in\agentset} \actions_{i}} \times \states} \to [0,1]
    }$ is the \emph{transition function}. It has the property that
    for all $s\in \states$, for all $(a_{1},a_{2},\dots,a_{\aN})\in
    \prod_{i\in\agentset}\actions_{i}$,
    $\sum_{s'\in\states} \transition(s, a_{1}, a_{2}, \dots, a_{\aN},
    s') = 1$.
  \item
    $\reward_{i}\colon \states \times \prod_{i\in\agentset} \actions_{i} \times \states \to \R$ is the
    \emph{reward function} for agent $i$.
  \item $\observations_{i}$ is the set of possible \emph{observations}
    for agent $i$.
  \item
    $\obsfunc_{i}\colon {{\actions_{i} \times \states \times
        \observations_{i}} \to [0,1]}$ is the \emph{observation
      function}.
    It has the property that $\sum_{\observation\in\observations_{i}}\obsfunc_{i}(a,s,\observation)=1$
    for all ${a\in\actions_{i}}$ and ${s\in\states}$. 
  \end{itemize}
\end{definition}

\subsection{Extensive Form Games}
\label{sec:efgdef}

The definition given here follows closely that of~\citet{osborne1994gametheory}, to which we refer the reader for a more in-depth discussion of Extensive Form Games and their formal definition.

\begin{definition}
\label{def:efg}
  An Extensive Form Game is defined by:
  \begin{itemize}
    \item A set of agents $\agentset = \set{1,2,\dots,\aN}$.
    \item A ``Nature'' player
    denoted as ``agent'' $0$. For convenience, we define
          $\agentenvset := \agentset \cup \set{0}$. The Nature agent is responsible for describing the random, stochastic, or luck-based elements of the game, as described below.
    \item A set $\actseq$ of \emph{action sequences}. An action sequence is a tuple $\vec{a} = (a_1, a_2, \dots, a_k)$ where each element indicates an action taken by an agent. In infinite games, action sequences need not be finite. The set $\actseq$ indicates all possible sequences of actions that may be taken in the game (i.e., ``histories'' of players' moves or agents' actions). It satisfies the following properties:
          \begin{itemize}
            \item The empty sequence is in the set: $\varnothing\in\actseq$.
            \item If $(a_{1}, \dots, a_{k}) \in \actseq$, then for
                  $l<k$ we also have
                  $(a_{1}, \dots, a_{l}) \in \actseq$.
            \item In infinite games, if an infinite sequence
                  $(a_{1}, a_{2}, \dots)$ satisfies the property that for all
                  $k$, $(a_{1}, a_{2}, \dots, a_{k}) \in \actseq$, then
                  $(a_{1}, a_{2}, \dots)\in \actseq$.
          \end{itemize}
          %
          For a finite sequence $\vec{a} = (a_{1}, \dots, a_{k})$, denote by $(\vec{a}, a)$ the sequence $(a_{1}, \dots, a_{k}, a)$. Then the set of actions available
          in the next turn following a sequence $\vec{a}$ is given
          by $\actions(\vec{a}) := \set{a \mid (\vec{a}, a) \in \actseq}$ (for convenience, we define $\actions(\vec{a}) = \varnothing$ if $\vec{a}$ is infinite). We say a sequence of actions $\vec{a}$ is \emph{terminal} if it is either infinite or if it is a maximal finite sequence, i.e.\ $\vec{a}$ is terminal if and only if $\actions(\vec{a}) = \varnothing$. We denote the set of terminal sequences by $\terminals := \set{\vec{a} \mid \actions(\vec{a}) = \varnothing}$.
    \item A function $\agentturn \colon (\actseq \setminus \terminals) \to \agentenvset$,
          which specifies the agent whose turn it is to act next after a given
          sequence of actions.
          Note that this is not stochastic, but random player order can be captured by inserting a Nature turn.
    \item A probability distribution $\transition(\vec{a}, \cdot)$ for Nature's actions. It is defined only for action sequences for which Nature acts next, i.e.\ sequences $\vec{a}\in\actseq$ for which
          $\agentturn(\vec{a}) = 0$. Specifically, $\transition(\vec{a}, a)$ is the probability that Nature takes
          action $a$ after the sequence of actions $\vec{a}$ has occurred.
    \item For each agent $i\in\agentset$, a \emph{partition} $\infopart_{i}$ of the sequences
          of actions
          $\actseq_{i} := \set{\vec{a} \mid \agentturn(\vec{a}) = i}$. The partition $\infopart_{i}$ is
          called the \emph{information partition} of agent $i$, and
          elements of $\infopart_{i}$ are called \emph{information sets}. For convenience, define
          $\infopart := \bigcup_{i\in\agentset} \infopart_{i}$.
          The information sets must obey the additional
          property that for any information set $h\in \infopart$ and any two action sequences
          $\vec{a}, \vec{a}'\in \infopart$, 
          we have $\agentturn(\vec{a}) = \agentturn(\vec{a}')$ and
          $\actions(\vec{a}) = \actions(\vec{a}')$.
    \item For each agent $i\in \agentset$, a \emph{reward function}
          $\reward_{i} \colon T\to \R$.
  \end{itemize}
\end{definition}

\subsection{Agent Environment Cycle Games}
\label{sec:aecdef}

As mentioned in \Cref{aec_model}, the stochastic nature of the state transitions is modeled as an ``environment'' agent, which does not take an action but rather transitions randomly from the current state to a new state according to some given probability distribution. With the stochasticity of state transitions separated out as a distinct ``environment'' agent, we can then model the transitions of the actual agents deterministically. To this end, each (non-environment) agent $i$ has a deterministic transition function $\dettrans_i$ which depends only on the current state and the action taken, while the environment has a stochastic transition function $\transition$ which transitions to a new state randomly depending on the current state (it may depend on the actions taken previously by the agents, since the current state is determined by these actions).

\begin{definition} \label{def:aec}
  An \emph{Agent-Environment Cycle Game} (AEC Game) is a tuple
  $\langle \states, s_0, \aN, (\actions_{i})_{i\in\agentset}
  ,
  (\dettrans_{i})_{i\in\agentset},
  \transition,
  (\rewardspace_{i})_{i\in\agentset},
  (\reward_{i})_{i\in\agentset},
  ,
  (\observations_{i})_{i\in\agentset},
  ,
  (\obsfunc_{i})_{i\in\agentset},
  ,
  \agtrans
  \rangle$, where:
  \begin{itemize}
  \item $\states$ is the set of possible \emph{states}.
  \item $s_0$ is the \emph{initial state}.
  \item $\aN$ is the \emph{number of agents}. The agents are numbered
    $1$ through $\aN$. There is also an additional ``environment''
    agent, denoted as agent $0$. We denote the set of agents along
    with the environment by $\agentenvset := \agentset \cup \set{0}$.
  \item $\actions_{i}$ is the set of possible \emph{actions} for agent
    $i$. For convenience, we further define
    $\actions_{0} = \set{\varnothing}$ (i.e., a single ``null action'' for environment steps) and
    $\actions := \bigcup_{i\in\agentenvset} \actions_{i}$. 
  \item
    $\dettrans_{i}\colon {{\states \times \actions_{i}} \to \states}$
    is the \emph{transition function for agents}. State
    transitions for agent actions are deterministic.
  \item
    $\transition\colon {{\states \times
        \states} \to [0,1] }$ is the \emph{transition
      function for the environment}. State transitions for environment steps are stochastic: $\transition(s, s')$ is the probability that the
    environment transitions into state $s'$ from state $s$.
  \item $\rewardspace_i \subseteq \R$ is the set of possible rewards for agent $i$. We assume this is \emph{finite}.
  \item
    $\reward_i \colon {{\states \times \agentenvset \times \actions
        \times \states \times \rewardspace_i} \to [0,1]}$ is the
    \emph{reward function} for agent $i$.
    $\rewardspace_{i}\subseteq \R$ denotes the set of all possible
    rewards for agent $i$ (which we assume to be finite). 

    $\reward_{i}$ is the \emph{reward function} for agent $i$. The set
    of all possible rewards for each agent is assumed to be finite,
    which we denote $\rewardspace_i\subseteq \R$.  It is
    \emph{stochastic}: $\reward_{i}(s, j, a, s', r)$ is the
    probability of agent $i$ receiving reward $r$ when agent $j$ takes
    action $a$ while in state $s$, and the game transitions to state
    $s'$.
    We also define
    $\rewardspace := \bigcup_{i\in\agentset} \rewardspace_i$.
  \item $\observations_{i}$ is the set of possible \emph{observations}
    for agent $i$.
  \item
    $\obsfunc_{i}\colon {{\states \times \observations_{i}} \to
      [0,1]}$ is the \emph{observation function} for agent
    $i$. $\obsfunc_{i}(s, \observation)$ is the probability of agent
    $i$ observing $\observation$ while in state $s$.
  \item
    $\agtrans \colon {{\states \times \agentenvset \times \actions \times \agentenvset} \to
      [0,1]}$ is the \emph{next agent} function. This means that
    $\agtrans(s, i, a, j)$ is the probability that agent $j$ will be
    the next agent permitted to act given that agent $i$ has just
    taken action $a$ in state $s$. This should attribute a non-zero probability only when $a\in \actions_i$.
  \end{itemize}
\end{definition}

In this definition, the game starts in state $s_0$ and the environment agent acts first. Having the environment agent act first allows the first actual agent to act to be determined randomly if desired (choosing the first agent deterministically can be done easily by having the environment simply do nothing in this first step). The game then
evolves in ``turns'' where in each turn an agent $i$ receives an
observation $\observation_{i}\in\observations_{i}$ (any given
observation $\observation$ is seen with probability
$\obsfunc_{i}(s, \observation)$) and, based on this observation,
chooses an action $a_{i}\in \actions_{i}$. The game then transitions
from the current state $s$ to a new state $s'$ according to the
transition function. If $i\in\agentset$, the state transition is
deterministically $\dettrans_{i}(s, a_{i})$. If $i=0$, the new state
is stochastic, so state $s'$ occurs with probability
$\transition(s, s')$. Then, a new agent $i'$ is determined according
to the ``next agent'' function, so that $i'$ is next to act with
probability $\agtrans(s, i, a_{i}, i')$. The observation
$\observation_i$ that is received is random, occurring with
probability $\obsfunc_i(s,\observation_i)$.  Note that we can allow
for the state to transition randomly in response to an agent's action
by simply inserting an ``environment step'' immediately following an
agent's action, by setting $\agtrans(s, i, a_{i}, 0)=1$ and allowing
the following environment step to transition the state randomly.  At
every step, every agent $j$ receives the partial reward $r'$ with
probability $\reward_j(s,i,a_i,s',r')$.

\section{Omitted Proofs}
\label{sec:omitted_proofs}

\subsection{POSGs are Equivalent to AEC Games}
\label{app:aec_to_posg}

The inclusion of the stochastic $\agtrans$ (next-agent) function
in the definition of AEC games
allows for capturing many turn-based games with complex turn orders
(consider Uno, for instance, where players may be skipped or the order reversed). It is not immediately
obvious that this allows for representing games in which agents act
simultaneously. 
However, we show here that in fact AEC games can be used to theoretically model games with simultaneous actions.

To see this, imagine simulating a POSG by way of a ``black box'' which takes the actions of all agents simultaneously, and then --- one by one --- feeds them to a purpose-built AEC game whose states are designed to ``encode'' each agent's action, ``queueing'' them up over the course of $\aN$ steps (one for each agent). Once all of the actions have been fed to the AEC game, a single environment step resolves these ``queued up'' actions all at once. If we design the AEC game in the right way, this total of $\aN+1$ steps ($\aN$ for queueing the actions, and one for the environment to resolve the joint action) produces an outcome that is identical to the result of a single step in the original POSG. This is formalized below.
\begin{theorem} \label{thm:posgtoaec}
  For every POSG, there is an equivalent AEC Game.
\end{theorem}
\begin{proof}[Proof of \Cref{thm:posgtoaec}]
  Let
  $G = \langle \states, \aN, \set{\actions_{i}}, \transition,
  \set{\reward_{i}}, \set{\observations_{i}}, \set{\obsfunc_{i}}
  \rangle$ be a POSG. To prove this, it will be necessary to show
  precisely what is meant by ``equivalent.'' We will construct a new
  AEC Game $G_{\mathrm{AEC}}$ in such a way that for every $\aN+1$
  steps of $G_{\mathrm{AEC}}$ the probability distribution over
  possible states is identical to the state distribution for $G$ after
  a single step, the distributions over observations received by each agent is identical in $G$ and in $G_{\mathrm{AEC}}$, and the reward obtained by each agent is the
  same.

  We define $G_{\mathrm{AEC}}$ as follows:
  \[
  G_{\mathrm{AEC}} = \langle \states', \aN, \set{\actions_{i}},
  \set{\dettrans_{i}}, \transition', \set{\reward'_{i}},
  \set{\observations_{i}}, \set{\obsfunc'_{i}}, \agtrans \rangle
  \]
  where
  \begin{itemize}
  \item
    $\states' = \states \times \actions_{1} \times \actions_{2} \times
    \dots \times \actions_{\aN}$. That is, an element of $\states'$
    is a tuple $(s, a_{1}, a_{2}, \dots, a_{\aN})$ where $s\in\states$
    and for each $i\in\agentset$, $a_{i}\in \actions_{i}$.
  \item
    ${\dettrans_{i}({(s, a_{1}, a_{2}, \dots, a_{i}, \dots, a_{\aN})},
      a'_{i})} = {(s, a_{1}, a_{2}, \dots, a'_{i}, \dots,
      a_{\aN})}$.
  \item For $\mathbf{s} = {(s, a_{1}, a_{2}, \dots, a_{\aN})}$ and
    $\mathbf{s'} = {(s', a_{1}, a_{2}, \dots, a_{\aN})}$, we define
    $\transition'(\mathbf{s}, \mathbf{s'}) = 
    \transition(s, a_{1}, a_{2}, \dots, a_{\aN}, s')$. 
    If $\mathbf{s}$ and $\mathbf{s'}$ are such that $a_{i} \ne a_{i}'$
    for any $i\in \agentset$, then
    $\transition'(\mathbf{s}, \mathbf{s'}) = 0$.
  \item For $\mathbf{s} = (s, a_{1}, a_{2}, \dots, a_{\aN})$,
    $\mathbf{s'} = (s', a_{1}, a_{2}, \dots, a_{\aN})$, and
    $\mathbf{r} = \reward_{i}(s, a_{1}, a_{2}, \dots, a_{\aN}, s')$,
    we let
    $\reward'_{i}(\mathbf{s}, 0, \varnothing, \mathbf{s'}, \mathbf{r})
    = 1$. We define $\reward'_{i} = 0$ for all other cases.
  \item
    $\obsfunc'_{i}(s, a_{1}, a_{2}, \dots, a_{\aN}) = \obsfunc_{i}(s)$
  \item
    $\agtrans((s, a_{1}, a_{2}, \dots, a_{\aN}), i, a_{i}', j) = 1$ if
    $j \equiv i+1 \pmod{\aN+1}$ (and equals $0$ otherwise). 
  \end{itemize}
  The AEC game $G_{\mathrm{AEC}}$ begins with agent $1$. If the
  initial state of the POSG $G$ was $s_{0}$, then the initial state of
  $G_{\mathrm{AEC}}$ is $(s_{0}, \cdot, \cdot, \dots, \cdot)$, where
  all but the first element of the tuple are chosen arbitrarily.  

  Let $P_{t,s}$ be the probability that the POSG $G$ is in state $s$
  after $t$ steps. For an action vector
  $\mathbf{a} = (a_{1}, \dots, a_{\aN}) \in \actions_{1}\times \dots
  \times \actions_{\aN}$, let $P'_{t,s,\mathbf{a}}$ be the probability
  that $G_{\mathrm{AEC}}$ is in state $(s, a_{1}, \dots, a_{\aN})$
  after $t$ steps. Finally, let
  $P'_{t,s} = \sum_{\mathbf{a}\in \actions_{1} \times \dots \times
    \actions_{\aN}} P'_{t,s,\mathbf{a}}$.

  Trivially, $P_{0,s} = P'_{0,s}$ for all $s\in \states$. Now, suppose
  that after $t$ steps of $G$, $P_{t,s} = P'_{t(\aN+1),s}$ for all
  $s\in\states$ (our inductive hypothesis). For any joint action
  $\mathbf{a} = (a_{1}, \dots, a_{\aN})$, the state distribution of
  $G$ at step $t+1$ if the joint action $\mathbf{a}$ is taken is given
  by
  $P_{t+1,s'} = P_{t,s}\cdot \transition(s, a_{1}, \dots, a_{\aN},
  s')$. Further, the reward obtained by agent $i$ for this joint
  action, if the new state is $s'$, is
  $\reward_{i}(s, a_{1}, \dots, a_{\aN}, s')$. Let
  $\mathbf{s} = (s, a_{1}, \dots, a_{\aN})$ and
  $\mathbf{s'} = (s', a_{1}, \dots, a_{\aN})$. Then, in
  $G_{\mathrm{AEC}}$, if the agents take actions
  $a_{1}, a_{2}, \dots, a_{\aN}$ respectively on their turns, the
  state distribution of $G_{\mathrm{AEC}}$ at step $(t+1)(\aN+1)$ is
  given by
  $P'_{(t+1)(\aN+1), s'} = P'_{(t+1)(\aN+1), s', \mathbf{a}} =
  P'_{t(\aN+1),s} \transition'(\mathbf{s}, \mathbf{s'})$. By the
  inductive hypothesis, $P'_{t(\aN+1),s} = P_{t,s}$, and by the
  definition of $\transition'(\mathbf{s}, \mathbf{s'})$ in
  $G_{\mathrm{AEC}}$, it is clear that
  $\transition'(\mathbf{s}, \mathbf{s'}) = \transition(s, a_{1},
  \dots, a_{\aN}, s')$. Thus,
  $P'_{(t+1)(\aN+1), s'} = P_{t,s} \transition(s, a_{1}, \dots,
  a_{\aN}, s') = P_{t+1, s'}$.  

  The above establishes a strict equivalence between the state
  distributions of $G$ at step $t$ and $G_{\mathrm{AEC}}$ at step
  $t(\aN+1)$ for any $t$. Between steps $t(\aN+1)+1$ and $(t+1)(\aN+1)$ of $G_{\mathrm{AEC}}$, each agent in turn receives an observation and then chooses its action. Specifically, agent $i$ acts at step $t(\aN)+i$ immediately after receiving an observation $\observation_{i}$ with probability $\obsfunc'_{i}(s, a_{1}, \dots, a_{\aN}) = \obsfunc_{i}(s)$. Thus, the marginal probability distribution (when conditioned on transitioning into state $s$) of the observation received by agent $i$ immediately after acting at time $t$ in $G$ is identical to the marginal distribution of the observation received by $i$ immediately before acting at time $t(\aN+1)+i$ in $G_{\mathrm{AEC}}$, i.e.\ $\Pr_{\mathrm{G},t}(\observation_i = \observation \mid s_{t}=s) = \Pr_{G_{\mathrm{AEC}},t(\aN+1)+i}(\observation_i=\observation \mid s_{t(\aN+1),0}= s)$.
  
  The second part of the equivalence is
  observing that the reward received by an agent $i$ in $G$ after the
  joint action $\mathbf{a}$ is taken is equivalent to the total reward
  received by agent $i$ in $G_{\mathrm{AEC}}$ across all steps from
  $t(\aN+1)+1$ through $(t+1)(\aN+1)$ when the agents take actions
  $a_{1}, \dots, a_{\aN}$ respectively.  We can see that this is
  indeed the case, since the rewards received by agent $i$ in
  $G_{\mathrm{AEC}}$ from step $t(\aN+1) + 1$ through step
  $(t+1)(\aN+1)$ is $0$ at every step but the environment step
  $(t+1)(\aN+1)$. By definition of $\reward'$ in $G_{\mathrm{AEC}}$,
  $\reward'_{i}(\mathbf{s}, 0, \varnothing, \mathbf{s'},
  \reward_{i}(s, a_{1}, \dots, a_{\aN}, s'))=1$, so the total reward
  received by any agent $i$ in $G_{\mathrm{AEC}}$ is
  $\reward_{i}(s, a_{1}, \dots, a_{\aN}, s')$. This establishes the
  second part of our equivalence (that the reward at step $t(\aN+1)$
  in $G_{\mathrm{AEC}}$ is identical to the reward at step $t$ of $G$,
  if the actions are the same).
\end{proof}


One way to think of this construction is that the actions are still resolved simultaneously via the \emph{environment step} (which is responsible for the stochastic state transition and the production of rewards); we simply break down the production of the joint action into smaller units whereby each agent chooses and ``locks in'' their actions one step at a time. A toy example to see this equivalence is to imagine a multiplayer card game in which each player has a hand of cards and each turn consists of all players choosing one card from their hand which is revealed simultaneously with all other players. An equivalent game has each player in sequence choosing a card and placing it face down on their turn, followed by a final action (the ``environment step'' in which all players simultaneously reveal their selected card.

At first, it may appear as though the AEC game is in fact \emph{more}
powerful than the POSG, since in addition to being able to handle
simultaneous-action games as shown above, it can represent sequential
games, including sequential games with complex and dynamic turn orders such as Uno (another aspect of our AEC
definition that seems more general than in POSGs is the fact that the
reward function in an AEC game is stochastic, allowing rewards to be
randomly determined).
However, it turns out that a POSG can be used to model a sequential
Handling the stochastic rewards and stochastic next-agent function is non-obvious and is omitted here due to space constraints; the construction and proof can be found in \autoref{app:aec_to_posg}.


We next show how to convert an AEC game to a POSG for the case of deterministic rewards.

\begin{definition}
  An AEC Game
  \[G = \langle \states, \aN, \set{\actions_i}, \set{\dettrans_i}, \transition, \set{\reward_i}, \set{\observations_i}, \set{\obsfunc_i}, \agtrans \rangle\]
  is said to have \emph{deterministic
    rewards} if for all $i,j\in\agentenvset$, all $a\in\actions_{j}$,
  and all $s,s'\in\states$, there exists a $\reward^{*}_{i}(s,j,a,s')$
  such that $\reward_{i}(s,j,a,s',r)=1$ for
  $r = \reward^{*}_{i}(s,j,a,s')$ (and 0 for all other $r$).
\end{definition}
Notice that an AEC Game with deterministic rewards may still depend on
the new state $s'$ which can itself be stochastic in the case of the
environment ($j=0$).

\begin{theorem}
  Every AEC Game with deterministic rewards has an equivalent POSG.
\end{theorem}
\begin{proof}
  Suppose we have an AEC game 
  \[
  G = \langle \states, \aN, \set{\actions_i}, \set{\dettrans_i}, \transition, \set{\reward_i}, \set{\observations_i}, \set{\obsfunc_i}, \agtrans \rangle
  \]
  with deterministic
  rewards. We define $G_{\mathrm{POSG}} = \langle \states', \aN, \set{\actions_i}, \transition', \set{\reward'_i}, \set{\observations_i}, \set{\obsfunc_i} \rangle$ as follows.
  \begin{itemize}
  \item $\states' = \states\times \agentenvset$
  \item
    $\transition'((s,i), a_{1}, \dots, a_{\aN}, (s',i')) =
    \agtrans(s, i, a_{i}, s', i') \cdot \Pr(s' 
    \mid s,i,a_{i})$, where
    \[
      \Pr(s' 
      \mid s,i,a_i) =
      \begin{cases}
        1 & \text{if } i>0, T(s,a_{i})=s' \\
        \transition(s,s') & \text{if } i=0 \\
        0 & \text{o/w} 
      \end{cases}
    \]
    \item $\reward'_i((s,j), a, (s',j')) = \reward^{*}_i(s, j, a, s')$
  \end{itemize}
  In this construction, the new state in the POSG encodes information about which agent is meant to act. State transitions in the POSG therefore encode both the state transition of the original AEC game and the transition for determining the next agent to act.  In each step, the state transition depends only on the agent who's turn it is to act (which is included as part of the state).
  
  This construction adapts POSGs to be strictly turn-based so that it is able to represent AEC Games.
\end{proof}

We now present the full proof.
\begin{theorem}
  Every AEC Game has an equivalent POSG.
\end{theorem}
\begin{proof}
  Suppose we have an AEC game
  $G = \langle \states, \aN, \set{\actions_i}, \set{\dettrans_i},
  \transition, \set{\reward_i}, \set{\observations_i},
  \set{\obsfunc_i}, \agtrans \rangle$, and
  $\rewardspace$ is the (finite) set of all possible rewards.  We
  define
  $G_{\mathrm{POSG}} = \langle \states', \aN, \set{\actions_i},
  \transition', \set{\reward'_i}, \set{\observations_i},
  \set{\obsfunc_i} \rangle$ as follows.
 
    The state set is $\states' = \states\times \agentenvset \times
    \rewardspace^{\aN}$. An element of $\states'$ is a tuple
    $(s, i, \mathbf{r})$, where
    $\mathbf{r} = (r_{1},r_{2},\dots,r_{\aN})$ is a vector of rewards
    for each agent.
    
    The transition function is given by 
    \[
    \begin{split}
    &\transition'((s,i,\mathbf{r}), a_{1}, a_{2}, \dots, a_{\aN}, (s',i',\mathbf{r'})) = \\
    &\qquad\agtrans(s, i, a_{i}, s', i') 
    \Pr(s' 
    \mid s,i,a_{i})
    \prod_{j\in\agentset}\reward_{j}(s, i, a_{i}, s', \mathbf{r'}_{i})
    \end{split}
    \]
    where
    \[
      \Pr(s' 
      \mid s,i,a_i) =
      \begin{cases}
        1 & \text{if } i>0 \text{ and } T(s,a_{i})=s' \\
        \transition(s,s') & \text{if } i=0 \\
        0 & \text{o/w} 
      \end{cases}
    \]
    
    The reward function is given by $\reward'_i((s,j,\mathbf{r}), a, (s',j',\mathbf{r'})) = \mathbf{r'}_{i}$
\end{proof}

\end{document}